\documentclass[letterpaper]{article} 
\usepackage[preprint]{aaai2027}  
\usepackage[hyphens]{url}  
\usepackage{graphicx} 
\usepackage{natbib}  
\usepackage{caption} 
\usepackage{booktabs}
\usepackage{multirow}
\usepackage{amsmath}
\usepackage{amssymb}
\newcommand{\codeurl}{\url{https://anonymous.4open.science/r/not-the-dimension-the-norm-DED6}}

\title{Not the Dimension, the Norm:\\What Matters in Gradient-Free Weight Perturbation of Language Models}

\author{
    Taeyeong Kim\textsuperscript{\rm 1},
    Ahhyun Kim\textsuperscript{\rm 1},
    TaeHyeon Kim\textsuperscript{\rm 1},
    Unggi Lee\textsuperscript{\rm 2}\corresponding
}
\affiliations{
    \textsuperscript{\rm 1}Chosun University\quad
    \textsuperscript{\rm 2}Korea University Sejong Campus\\
    codingchild@korea.ac.kr
}

\begin{document}

\maketitle

\begin{abstract}
Adapting a language model to a task no longer requires training all of its weights, and a line of parameter-efficient methods has driven the trainable count from billions down to a handful of scalars. Gradient-free adaptation, which samples random weight perturbations and keeps the ones that score well, has not followed that trajectory and still perturbs every entry of the weight tensor. It is unknown whether that full-weight search is necessary, and more fundamentally which property of a perturbation makes it work at all, because existing methods vary the search space, the perturbation scale, and the aggregation together. We resolve this by intervening on one factor at a time inside a fixed pipeline, holding candidate scoring and voting constant while we vary the search dimension, the subspace that carries the perturbation, and its norm. Perturbing a frozen frame of 12 to 16 scalars stays 1.8 accuracy points behind full-weight search on average across 49 model-benchmark cells, trailing it in 36 of them. Neither the dimension nor the choice of basis explains that performance. A random frame whose Grassmann overlap with the SVD frame is at chance level performs identically once a single scale factor is matched, and at large scales the SVD directions collapse first. What survives is the perturbation norm, whose usable range closes within a factor of five across seven models and stays flat inside. The perturbation norm is therefore the one factor with a failure mode, and its safe region transfers across scale and family. The design question narrows from \emph{which subspace to perturb} to \emph{how hard to shake}.
\end{abstract}

\section{Introduction}
\label{sec:intro}

Adapting a pretrained language model to a task is no longer a matter of updating all of its weights. Parameter-efficient methods have driven the trainable count from billions to millions, then to thousands, and in the extreme to about a dozen scalars per model \citep{hu2022lora, balazy2025loraxs, morris2026learningreason13parameters}, and they work because the update that a task requires lives in a very small subspace. Gradient-free adaptation reaches the same goal without any gradient at all. It samples random perturbations of the frozen weights, scores each candidate on a small labeled set, and aggregates the survivors, which makes adaptation possible wherever backpropagation is impractical \citep{salimans2017evolutionstrategiesscalablealternative, mehrtash2020pep, gan2026thickets}.\footnote{Code, configurations, and per-cell results are available at \codeurl.}

\begin{figure}[t]
\centering
\includegraphics[width=0.99\columnwidth]{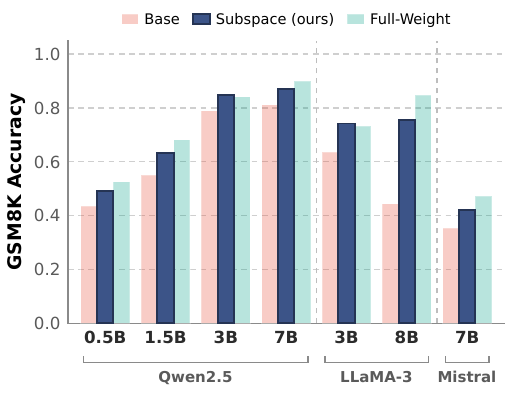}
\caption{Subspace search stays close to full-weight search across models. GSM8K accuracy for all seven models, comparing the unperturbed model, a frozen frame of 12 to 16 scalars, and full-weight RandOpt. The full grid is in Table~\ref{tab:full-grid}.}
\label{fig:scale}
\end{figure}

\begin{figure*}[t]
\centering
\includegraphics[width=0.90\textwidth]{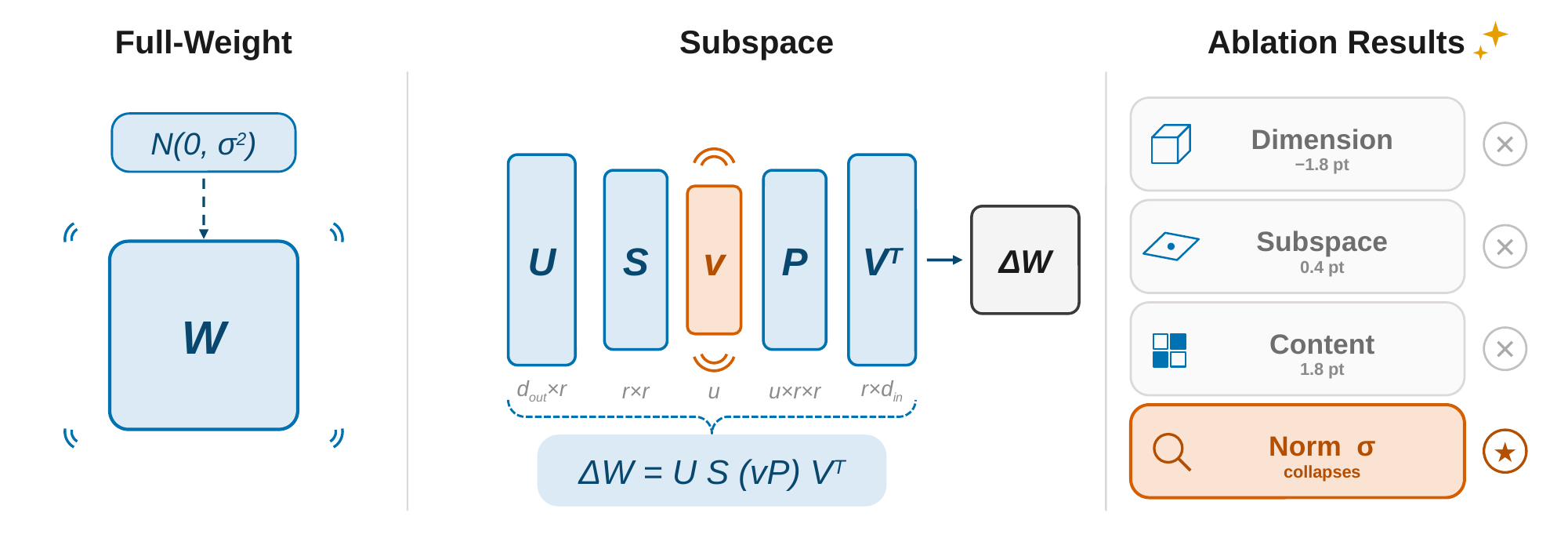}
\caption{The testbed and the verdicts. \emph{Left}. Gradient-free search perturbs every entry of the weight tensor. \emph{Middle}. Our testbed perturbs only the small vector $v$ inside a frozen frame and reaches the same accuracy, with scoring and voting held fixed. \emph{Right}. Each entry gives how far accuracy moves when that factor is misspecified, as a 49-cell mean for dimension and a control-cell spread for the other two. Only the perturbation norm collapses the model outside its safe window.}
\label{fig:overview}
\end{figure*}

The gradient-free branch, however, never inherited the low-dimensional lesson. Its standard recipe still perturbs every entry of the weight tensor, hundreds of millions of parameters at a time, and recent work builds on that full-weight regime rather than questioning it \citep{gan2026thickets}. The asymmetry is hard to justify. If adaptation is low-dimensional when we \emph{learn} the update, it is unclear why the full space would be needed when we merely \emph{sample} it. The deeper problem is that no one knows which property of a random perturbation makes it succeed, because existing methods change the search space, the perturbation scale, and the aggregation strategy together, so their contributions cannot be separated.

We settle both questions by intervening on one factor at a time inside a fixed pipeline. Candidate scoring and majority voting are held to their standard form throughout, and we vary only the geometry upstream of them, namely the search dimension, the frame that carries the perturbation, the information inside that frame, and the perturbation norm. Causal interventions run on a control cell with three seeds, and generality is verified separately on a grid of seven models from 0.5B to 8B parameters across three families and seven benchmarks spanning math reasoning, arithmetic puzzles, code generation, commonsense, and chemistry.

The eliminations are almost entirely negative. Perturbing a frozen frame of twelve to sixteen scalars recovers most of full-weight search at every scale we test, trailing it by 1.8 points on average and by 9.1 at worst (Figure~\ref{fig:scale}), and enlarging that frame by two orders of magnitude leaves that distance unchanged, so dimension does not govern performance. Neither does the subspace. A random frame whose Grassmann overlap with the SVD frame is at the chance level performs identically once a single scale factor is matched, replacing the singular values or vectors changes nothing, and at large scales the SVD directions collapse before random ones. What survives is the perturbation norm, whose usable range closes within a factor of five across all seven models and is flat enough inside that a fixed value costs 0.1 points.

Our study replaces the prevailing intuition that gradient-free adaptation works by searching a large or well-chosen weight subspace with a sharper account in which a single scalar, the perturbation norm, governs the outcome and requires no tuning.

Our contributions are threefold.
\begin{itemize}
    \item We show that a frozen frame of a dozen scalars recovers full-weight gradient-free search to within 1.8 accuracy points on average, and 9.1 at worst, while searching seven orders of magnitude fewer parameters.
    \item We eliminate the subspace as an explanatory factor and identify what the SVD frame actually supplies, which is automatic per-layer calibration of the perturbation scale.
    \item We show that the perturbation norm is the only live factor and that its usable range transfers across model sizes and families without search, yielding a tuning-free prescription.
\end{itemize}

\section{Related Work}
\label{sec:related}

\subsection{Low-Dimensional Adaptation}
The trajectory by which prior work has compressed the trainable update of a pretrained model is what motivates our question. It begins with intrinsic-dimension measurements, which optimize inside a randomly oriented low-dimensional subspace and find that few coordinates suffice \citep{li2018intrinsic, aghajanyan2021intrinsic}. Low-rank adapters reduce the update to a trainable product of two thin matrices \citep{hu2022lora}, and later variants initialize or allocate that product using the pretrained weight's own spectrum \citep{meng2024pissa, zhang2023adalora}. A further family freezes the projection entirely and trains only a small inner matrix or a per-layer scaling vector inside it \citep{balazy2025loraxs, kopiczko2024vera}. Pushed to its extreme, this line trains roughly a dozen scalars per model \citep{morris2026learningreason13parameters}, and recent work reports that the frozen scaffold carrying those scalars can be random without loss \citep{hazan2026littlerankgoeslong}. All of these methods keep a gradient backend, so what a frozen frame contributes when no gradient is available remains open, and that is the regime we study.

\subsection{Gradient-Free Perturbation}
Gradient-free methods adapt a model by sampling perturbations rather than differentiating through it. Evolution strategies and plain random search treat the weights as a black box and update them from scored samples \citep{salimans2017evolutionstrategiesscalablealternative, mania2018simple}, a line in which step size has long been the dominant control, to the point of requiring online adaptation \citep{hansen2001derandomized}. Zeroth-order methods instead use paired perturbations to estimate a gradient from forward passes alone \citep{malladi2023mezo}, while perturbation ensembles skip parameter updates altogether by aggregating the predictions of several perturbed copies \citep{mehrtash2020pep, wang2023selfconsistency}. The recipe we adopt as a testbed, RandOpt \citep{gan2026thickets}, adds train-set scoring and top-$K$ aggregation over sampled full-weight perturbations, motivated by the observation that task experts are dense around pretrained weights. Across this literature these three axes are chosen jointly and reported as a package, which is what our interventions separate. To relate perturbation frames we borrow the Grassmann subspace comparison introduced for adapters \citep{hu2022lora} and the reading of singular values as directional output sensitivity established for generative latent spaces \citep{park2023understanding, liang2026blessing}. The latter argues high-dimensional fine-tuning is advantageous, which our grid does not reproduce when the update is sampled rather than learned.

\section{Experimental Setup}
\label{sec:framework}

\paragraph{Testbed.}
Our testbed is gradient-free random search in a frozen subspace (Figure~\ref{fig:overview}, middle). Following TinyLoRA \citep{morris2026learningreason13parameters}, each target layer of a frozen model is decomposed by a top-$r$ singular value decomposition (SVD), $W = U S V^\top$, and perturbed in place as
\begin{equation}
W' \;=\; W + \underbrace{U\, S(vP)\, V^\top}_{\Delta W},
\label{eq:testbed}
\end{equation}
where $W \in \mathbb{R}^{d_\text{out} \times d_\text{in}}$ is a frozen weight of one target linear layer and $U \in \mathbb{R}^{d_\text{out} \times r}$, $S \in \mathbb{R}^{r \times r}$, $V^\top \in \mathbb{R}^{r \times d_\text{in}}$ come from its truncated SVD at rank $r{=}2$. The vector $v \sim \mathcal{N}(0, \sigma^2 I)$, $v \in \mathbb{R}^{v_\text{dim}}$, is the only perturbed quantity ($v_\text{dim}{\approx}12$ scalars for the whole model, tied across modules), $P \in \mathbb{R}^{u \times r \times r}$ is a frozen projection whose $u$ slices contract with the $u$ entries of $v$ that this module reads, $u$ being that module's share of the tied budget, giving $vP \in \mathbb{R}^{r \times r}$, and $U$, $S$, $V$ are frozen. There is no gradient and no optimizer. One adaptation run follows the RandOpt protocol \citep{gan2026thickets}, which samples $N$ candidate vectors $v$, scores each on a small train set, and aggregates the answers of the top-$K$ candidates by majority vote.

\paragraph{Fixed protocol, isolated dials.}
The scoring--selection--voting pipeline is held fixed in its standard form throughout; our interventions target only the geometry upstream of it. We decompose the testbed into a \emph{structural} axis (which frame lifts $v$ to $\Delta W$) and a \emph{numeric} axis (perturbation scale $\sigma$, population $N$, rank $r$, scoring-set size $n_\text{fit}$, vote size $K$), and change exactly one of them at a time. In prior methods these axes are entangled, which is why it has been unclear which of them produces the performance. Section~\ref{sec:dimension} sweeps the numeric axis and Section~\ref{sec:subspace} replaces the structural one.

\paragraph{Two-tier design.}
Causal interventions are expensive, so we run them on a control cell (Qwen2.5-0.5B-Instruct on GSM8K, three seeds) and verify generality separately on the $7{\times}7$ grid. The capacity sweep and the $\sigma$ window are additionally verified at grid scale; the frame and content interventions are control-cell results. Models span Qwen2.5 (0.5B/1.5B/3B/7B) \citep{qwen2024qwen25}, LLaMA-3.2-3B, LLaMA-3.1-8B, and Mistral-7B-v0.1. Benchmarks cover GSM8K \citep{cobbe2021gsm8k}, MATH-500 \citep{hendrycks2021math}, OlympiadBench \citep{he2024olympiadbench}, Countdown \citep{gan2026thickets}, MBPP \citep{austin2021mbpp}, ROCStories \citep{mostafazadeh2016rocstories}, and USPTO-50K \citep{schwaller2019molecular}; splits are listed in Table~\ref{tab:splits}. All evaluations use the instruct chat template and greedy decoding.

\paragraph{Setup.}
Unless stated otherwise, each cell runs population $N{=}1{,}000$ over the $\sigma$ grid $\{0.01, 0.05, 0.1, 0.5, 1.0\}$, scores candidates on 200 training examples, and reports the best ensemble over $K \in \{1, 10, 40, 50, 100\}$. Candidates are ranked by train score alone; only $K$ is chosen on the test set, under the same rule for every arm. Subspace capacity ranges over five configurations from \emph{utiny} (4 scalars) to \emph{large} (768--1{,}152 scalars); the full-weight RandOpt baseline shares the population, top-$K$ selection, and majority vote, replacing only the lift with Gaussian noise on every entry of $W$ on its own scale grid. Models are served in bf16 with vLLM \citep{kwon2023vllm} (implementation details in Appendix~H).

\begin{table}[!t]
\centering
\small
\begin{tabular}{lrrr}
\toprule
\textbf{Benchmark} & \textbf{Train} & \textbf{Test} & \textbf{Max tok.} \\
\midrule
GSM8K         & 200 & 1{,}319  & 1024 \\
MATH-500      & 200 & 300      & 2048 \\
OlympiadBench & 200 & 474      & 2048 \\
Countdown     & 200 & 300      & 1024 \\
MBPP          & 200 & 764      & 2048 \\
ROCStories    & 200 & 19{,}633 & 64   \\
USPTO-50K     & 200 & 4{,}902  & 64   \\
\bottomrule
\end{tabular}
\caption{Benchmark splits used in this work.}
\label{tab:splits}
\end{table}

\section{Do We Need to Perturb Everything?}
\label{sec:dimension}

\begin{figure}[!t]
\centering
\includegraphics[width=0.92\columnwidth]{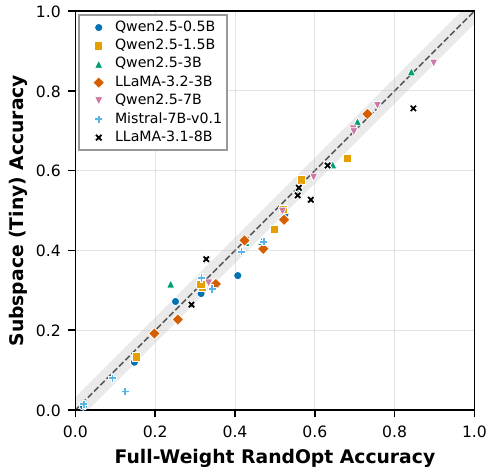}
\caption{Per-cell accuracy of subspace search (tiny, 12--16 parameters) versus full-weight RandOpt across the 49 model--benchmark grid. The shaded band is the median 95\% binomial confidence interval ($\pm$3.2 points). Full grid in Table~\ref{tab:full-grid}.}
\label{fig:parity}
\end{figure}

\paragraph{A 12-dimensional slice matches the full space.}
Our first question is whether the full weight space is necessary at all. Across the 49-cell grid, perturbing only the low-dimensional frame comes close to full-weight random search without reaching it. Full-weight search holds a small but consistent edge, 1.8 points on average, and the subspace trails it in 36 of the 49 cells (Table~\ref{tab:parity}, Figure~\ref{fig:parity}). The edge is real rather than noise, with a paired $t{=}-4.09$, a 95\% interval on the mean of $[-2.65, -0.93]$, and a sign test $p{=}3.5\times10^{-4}$. What the grid rules out is a large gap, not a gap, since per-cell binomial intervals run from $\pm$0.6 to $\pm$5.7 points. Restricting to the 35 cells outside the two saturated task families raises the mean gap to 2.3 points, so the comparison is not carried by those cells. Parity holds as scale grows within the Qwen2.5 family (Figure~\ref{fig:scale}), with no size trend in the per-model means (Table~\ref{tab:parity}).

Table~\ref{tab:full-grid} reports the full grid. Two task families require care in reading it. USPTO-50K and ROCStories impose strong output-format constraints under which many methods saturate at identical values, so their cells carry little signal about the search itself; and on Qwen2.5-0.5B\,/\,Countdown the base model cannot solve the task at all, so every method is uniformly low. The remaining task families, where the base model has traction, carry the comparison, and there the pattern is consistent. Wherever full-weight search gains over base, the subspace gains almost as much, and wherever full-weight search fails to move the needle, so does the subspace. Adaptation gains also grow with model scale on reasoning-heavy benchmarks, most visibly on Countdown, where both arms move from near zero at 0.5B to tens of points at 3B and above. This is consistent with the density argument of Neural Thickets \citep{gan2026thickets}, in which better-pretrained weights are surrounded by richer clouds of task-specialized solutions (Appendices~J and~K).

\begin{table}[!t]
\centering
\small
\begin{tabular}{lcc}
\toprule
\textbf{Model} & \textbf{$\Delta$ (pts)} & \textbf{95\% CI half-width} \\
\midrule
Qwen2.5-0.5B    & $-2.05$ & $\pm2.5$ \\
Qwen2.5-1.5B    & $-1.94$ & $\pm3.1$ \\
Qwen2.5-3B      & $+0.11$ & $\pm3.2$ \\
Qwen2.5-7B      & $-0.94$ & $\pm3.1$ \\
LLaMA-3.2-3B    & $-2.49$ & $\pm3.1$ \\
Mistral-7B-v0.1 & $-2.71$ & $\pm2.1$ \\
LLaMA-3.1-8B    & $-2.48$ & $\pm3.3$ \\
\midrule
All (49 cells)  & $-1.8$  & median $\pm3.2$ \\
\bottomrule
\end{tabular}
\caption{Per-model mean gap (subspace tiny $-$ full-weight). The right column gives single-arm binomial widths, which set the scale at which one cell can be read; the paired test of the difference is in the text.}
\label{tab:parity}
\end{table}

\begin{table*}[!t]
\centering
\scriptsize
\setlength{\tabcolsep}{4pt}
\renewcommand{\arraystretch}{0.90}
\begin{tabular}{ll|c|ccccc|c}
\toprule
\multirow{2}{*}{\textbf{Model}} & \multirow{2}{*}{\textbf{Benchmark}} & \textbf{Base} & \multicolumn{5}{c|}{\textbf{Subspace (ours)}} & \textbf{RandOpt} \\
 &  & (unperturbed) & utiny (4) & tiny (12-16) & small (40-64) & medium (192-288) & large (768-1152) & Full-Weight \\
\midrule
\multirow{7}{*}{Qwen2.5-0.5B}
 & GSM8K         & .434 & .475 & .491 & .486 & .483 & \underline{.504} & \textbf{.525} \\
 & MATH-500      & .313 & .367 & .337 & .353 & .340 & \underline{.390} & \textbf{.407} \\
 & OlympiadBench & .108 & \underline{.139} & .120 & .118 & .118 & .118 & \textbf{.148} \\
 & Countdown     & .044 & .007 & .010 & \textbf{.060} & \underline{.050} & \underline{.050} & .020 \\
 & MBPP          & .270 & .270 & \underline{.272} & \textbf{.277} & .268 & .264 & .251 \\
 & ROCStories    & .319 & .318 & .318 & .318 & \underline{.320} & \textbf{.321} & .318 \\
 & USPTO-50K     & .315 & .304 & .292 & .315 & \underline{.316} & \textbf{.322} & .315 \\
\midrule
\multirow{7}{*}{Qwen2.5-1.5B}
 & GSM8K         & .549 & .605 & .631 & \underline{.643} & .638 & .632 & \textbf{.682} \\
 & MATH-500      & .453 & .553 & \underline{.577} & \textbf{.580} & .527 & .553 & .567 \\
 & OlympiadBench & .215 & .302 & \underline{.308} & .289 & .298 & .306 & \textbf{.319} \\
 & Countdown     & .063 & .113 & \underline{.133} & .103 & .077 & .100 & \textbf{.153} \\
 & MBPP          & .500 & .507 & .503 & \textbf{.531} & \underline{.530} & .517 & .521 \\
 & ROCStories    & .466 & .459 & .453 & .465 & \underline{.477} & .402 & \textbf{.499} \\
 & USPTO-50K     & .073 & \textbf{.315} & \textbf{.315} & \textbf{.315} & \underline{.313} & \textbf{.315} & \textbf{.315} \\
\midrule
\multirow{7}{*}{Qwen2.5-3B}
 & GSM8K         & .790 & .846 & \textbf{.848} & \underline{.847} & .837 & .789 & .842 \\
 & MATH-500      & .603 & \textbf{.740} & \underline{.723} & .717 & .707 & .690 & .707 \\
 & OlympiadBench & .344 & \underline{.428} & .420 & \underline{.428} & \textbf{.430} & .416 & \underline{.428} \\
 & Countdown     & .084 & \underline{.460} & .423 & \textbf{.467} & .437 & .407 & \textbf{.467} \\
 & MBPP          & .605 & .619 & .615 & .623 & \underline{.636} & .623 & \textbf{.645} \\
 & ROCStories    & .538 & .540 & \underline{.547} & .538 & .497 & .386 & \textbf{.557} \\
 & USPTO-50K     & .063 & \underline{.315} & \textbf{.316} & \underline{.315} & \underline{.315} & \underline{.315} & .239 \\
\midrule
\multirow{7}{*}{LLaMA-3.2-3B}
 & GSM8K         & .635 & .721 & .742 & \textbf{.755} & \underline{.750} & .738 & .732 \\
 & MATH-500      & .310 & .157 & .227 & \textbf{.330} & .207 & .243 & \underline{.257} \\
 & OlympiadBench & .125 & .169 & \underline{.192} & .181 & .186 & .188 & \textbf{.198} \\
 & Countdown     & .044 & .387 & .477 & \underline{.500} & .483 & .467 & \textbf{.523} \\
 & MBPP          & .429 & .470 & .404 & .462 & .384 & \textbf{.495} & \underline{.471} \\
 & ROCStories    & .414 & .418 & .425 & \underline{.442} & \textbf{.447} & .435 & .424 \\
 & USPTO-50K     & .178 & \underline{.335} & .316 & .325 & .255 & .327 & \textbf{.352} \\
\midrule
\multirow{7}{*}{Qwen2.5-7B}
 & GSM8K         & .811 & .839 & .869 & \underline{.889} & .857 & .861 & \textbf{.898} \\
 & MATH-500      & .703 & .750 & \underline{.763} & \textbf{.777} & .757 & .753 & .757 \\
 & OlympiadBench & .397 & .487 & .498 & .500 & .496 & \underline{.508} & \textbf{.519} \\
 & Countdown     & .292 & .563 & \underline{.583} & .577 & \underline{.583} & .567 & \textbf{.597} \\
 & MBPP          & .687 & .673 & \textbf{.704} & .691 & .685 & .692 & \underline{.696} \\
 & ROCStories    & .684 & .685 & \underline{.697} & .691 & .688 & .681 & \textbf{.698} \\
 & USPTO-50K     & .266 & .326 & .318 & .296 & \textbf{.367} & .320 & \underline{.334} \\
\midrule
\multirow{7}{*}{Mistral-7B-v0.1}
 & GSM8K         & .354 & .424 & .422 & \underline{.447} & .412 & .422 & \textbf{.472} \\
 & MATH-500      & .050 & .077 & \underline{.080} & .067 & .067 & .067 & \textbf{.093} \\
 & OlympiadBench & .013 & .015 & .015 & \textbf{.021} & \textbf{.021} & \underline{.019} & \textbf{.021} \\
 & Countdown     & .096 & .010 & .047 & .059 & \underline{.066} & .046 & \textbf{.125} \\
 & MBPP          & .296 & \underline{.319} & .304 & .305 & .312 & .310 & \textbf{.343} \\
 & ROCStories    & .343 & \underline{.416} & .397 & .398 & .394 & \textbf{.449} & \underline{.416} \\
 & USPTO-50K     & .271 & \underline{.323} & \textbf{.331} & .315 & \underline{.323} & .315 & .316 \\
\midrule
\multirow{7}{*}{LLaMA-3.1-8B}
 & GSM8K         & .444 & .691 & .756 & .821 & \underline{.834} & .803 & \textbf{.847} \\
 & MATH-500      & .330 & \textbf{.587} & .557 & .530 & .533 & .520 & \underline{.560} \\
 & OlympiadBench & .162 & .253 & .264 & .253 & .279 & \textbf{.295} & \underline{.291} \\
 & Countdown     & .514 & \underline{.541} & .538 & .520 & .534 & .530 & \textbf{.557} \\
 & MBPP          & .564 & .592 & .613 & .615 & .617 & \underline{.630} & \textbf{.632} \\
 & ROCStories    & .052 & .260 & .527 & .430 & .520 & \underline{.563} & \textbf{.590} \\
 & USPTO-50K     & .013 & .317 & \textbf{.378} & .291 & .315 & .315 & \underline{.328} \\
\bottomrule
\end{tabular}
\caption{Per-cell best ensemble accuracy across the 7$\times$7 model--benchmark grid for \emph{Base} (unperturbed), the five subspace capacities, and Full-Weight RandOpt. Despite searching $10^{5}$--$10^{9}$ times fewer parameters, the best of the five subspace capacities wins or ties on 25 of 49 cells against 27 for Full-Weight RandOpt, though that takes the best of five against one; the \emph{tiny} column alone wins or ties on 13 of 49, with a median absolute gap below two points. Per-row \textbf{best} and \underline{second-best} are marked; per-model exact $v_\text{dim}$ is in Appendix~B. USPTO-50K and ROCStories saturate under format constraints; Qwen2.5-0.5B\,/\,Countdown is unsolved by the base model.}
\label{tab:full-grid}
\end{table*}

\paragraph{Dimension is a dead dial.}
Parity between 12 dimensions and $10^{8}$ dimensions might still hide an effect somewhere in between, so we sweep the capacity explicitly. Within the same frame, the grid's capacity columns run from 4 to over 1{,}100 scalars, more than two orders of magnitude, and no capacity trend survives (Table~\ref{tab:full-grid}). The same paired test on \emph{utiny} against \emph{large} gives $t{=}-1.19$ and $[-2.66, +0.65]$, so the sweep excludes capacity effects above about 2.7 points rather than establishing flatness. Location is the one exception. Restricting to a layer group or module type costs 3.9 to 6.8 points, more than dimension does, though it lowers $v_\text{dim}$ at the same time (Appendix~C). The same holds for the remaining numeric dials. A one-axis-at-a-time sweep on the control cell over vote size $K$, population $N$, scoring-set size $n_\text{fit}$, and rank $r$ lands all ten of those variants within 3.9 points of the baseline (.491), which none of them exceeds by more than 0.1 points, with no single axis dominating. Per-$K$ breakdowns are in Appendix~C, where the baseline sits at or near the top at every ensemble size.

\paragraph{Neither does dimension buy or cost compute.}
One might at least expect the search dimension to matter for cost. It does not, for a structural reason. Profiling one adaptation run shows that scoring the candidates on the 200-example train set consumes 99.7\% of wall-clock time (about 3.2\,h for $N{=}1{,}000$ candidates on a single GPU), while applying a perturbation takes tens of milliseconds regardless of the frame. The search dimension simply has no lever on where the time goes; the subspace and full-weight pipelines cost the same within noise (full cost anatomy in Appendix~H).

\paragraph{Conclusion.}
Dimension is neither where the performance comes from nor where the cost goes. What determines adaptation performance must lie elsewhere.

\section{Does It Have to Be the SVD Subspace?}
\label{sec:subspace}

\paragraph{Replacing the frame changes nothing.}
If dimension is not the answer, perhaps the \emph{specific} subspace is. The frame of Eq.~\ref{eq:testbed} is derived from the weight's own SVD, and one might expect its principal directions to encode where useful task variation lives. We test this by replacing the structural axis piece by piece, holding the scalar count fixed at $v_\text{dim}{=}12$ so that random search always operates over the same space. Variant R1 removes the projection $vP$; R2 replaces the entire SVD frame with a raw Gaussian matrix; R3 further removes the projection, leaving the minimal lift $\Delta W = v \cdot M$ (formal definitions in Appendix~D). The only quantity matched across variants is the perturbation scale, via a single scalar per module, which we call $\alpha$, computed from weight-norm ratios. Under this norm matching, all four variants tie within 0.4 points (Table~\ref{tab:frame}), comparable to the per-variant seed spread of $\pm$0.4 to $\pm$0.7 points in that table. A frame carrying the full SVD prior and a frame carrying nothing but Gaussian noise are, to the search, the same frame.

\begin{table}[!t]
\centering
\small
\begin{tabular}{lccc}
\toprule
\textbf{Variant} & \textbf{SVD} & \textbf{$vP$} & \textbf{Acc.} \\
\midrule
Base            & \checkmark & \checkmark & .484 $\pm$ .007 \\
R1 (no $vP$)    & \checkmark & --         & .486 $\pm$ .006 \\
R2 (random matrix) & --      & \checkmark & .487 $\pm$ .006 \\
R3 (no $P$, $M$-only) & --   & --         & .488 $\pm$ .004 \\
\bottomrule
\end{tabular}
\caption{Frame ablation (norm-matched, $K{=}40$, three seeds). Replacing the SVD frame, or removing the projection entirely, matches Base within 0.4 points once the perturbation norm is matched.}
\label{tab:frame}
\end{table}

\paragraph{Nor does the content inside the frame matter.}
Wholesale replacement might conceivably destroy structure and compensating information at once, so we also replace the SVD's content \emph{surgically}, one component at a time while preserving the algebraic structure of the lift. Swapping the singular values for random diagonals, the singular vectors for random orthogonal bases, or the projection for random matrices, individually or in combination, moves accuracy by at most 1.8 points (Table~\ref{tab:swap}). Whatever information the decomposition stores about the pretrained weight, the search is not consuming it.

\begin{table}[!t]
\centering
\small
\setlength{\tabcolsep}{4pt}
\begin{tabular}{llll}
\toprule
\textbf{ID} & \textbf{Replaced} & \textbf{Replacement} & \textbf{Acc.} \\
\midrule
Base & --      & --                    & .491 \\
O1   & $S$     & diag-rand             & .488 $\pm$ .002 \\
O2   & $S$     & diag-uniform-mean     & .500 $\pm$ .007 \\
O3   & $U,V$   & orth-rand             & .497 $\pm$ .007 \\
O4   & $U,V,S$ & orth-rand + diag-rand & .473 $\pm$ .004 \\
O5   & $U,V,S$ & orth-rand + diag-unif & .496 $\pm$ .001 \\
P1   & $P$     & orth-rand             & .491 $\pm$ .002 \\
P2   & $P$     & diag-rand             & .494 $\pm$ .005 \\
\bottomrule
\end{tabular}
\caption{Content swap (best $K$, three seeds). Structure-preserving random replacement of $S$, $U/V$, and $P$ moves accuracy by at most 1.8 points from Base. Base is .491 here against .484 in Table~\ref{tab:frame}, which fixes $K{=}40$.}
\label{tab:swap}
\end{table}

\paragraph{The random frame is not secretly the SVD frame.}
A natural objection is that in high dimension a random frame might incidentally overlap the top singular directions. It does not. Following the precedent of measuring subspace similarity by principal angles \citep{hu2022lora} and representation geometry by geodesic distance \citep{park2023understanding}, we measure the Grassmann geodesic distance between frames. The SVD frame and a random orthogonal frame sit at 97\% of the theoretical maximum distance, with mean squared cosine of the principal angles $\cos^2\theta = 0.0057$, indistinguishable from the chance level $r/m$ for a rank-$r$ frame in an $m$-dimensional module, $r/m = 0.0055$ (Table~\ref{tab:grassmann}). The agreement is not an artifact of averaging, since each of the 168 target weight matrices sits at its own chance level, from $.0164$ against $.0156$ for the narrowest modules to $.00044$ against $.00041$ for the widest, and what residual spread remains tracks module width rather than depth. Two subspaces that share practically nothing produce identical performance.

\begin{table}[!t]
\centering
\small
\setlength{\tabcolsep}{3pt}
\begin{tabular}{lcc}
\toprule
\textbf{Comparison} & \textbf{$D_\text{geo}/D_\text{max}$ (min)} & \textbf{$\cos^2\theta$} \\
\midrule
SVD top-2 vs.\ random orth.\ 2-frame & 0.968 (0.867) & .0057 \\
SVD top-2 vs.\ dense $M$ top-2       & 0.970 (0.878) & --    \\
Chance level ($r/m$)                 & --            & .0055 \\
\bottomrule
\end{tabular}
\caption{Grassmann geodesic distance between the SVD frame and random frames, averaged over the 168 target modules with the per-module minimum in parentheses. Overlap matches the chance level.}
\label{tab:grassmann}
\end{table}

\begin{figure}[!t]
\centering
\includegraphics[width=0.95\columnwidth]{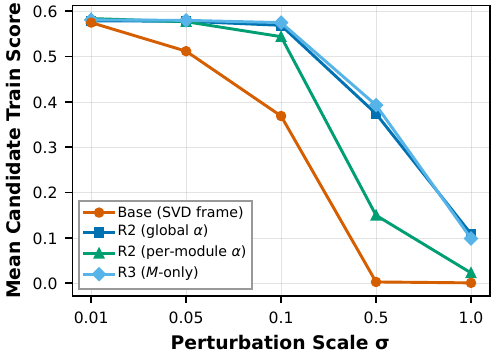}
\caption{Per-frame collapse as $\sigma$ grows, measured by mean candidate train score. All curves except the global-$\alpha$ control use per-module norm matching. The SVD frame collapses first, because its top singular directions are the highest-gain directions of each layer, so the function is shaken hardest at equal norm.}
\label{fig:collapse}
\end{figure}

\paragraph{At extreme scales, SVD directions are a liability.}
Direction indifference holds inside the moderate-scale regime; pushing $\sigma$ to extremes reveals a difference, and in the opposite direction from what the SVD prior would suggest. With per-module perturbation norms matched, the SVD frame collapses \emph{first} as $\sigma$ grows (Figure~\ref{fig:collapse}), because its top singular directions are each layer's highest-gain directions and at equal norm they shake the function hardest. Across both regimes we probe, the SVD directions never help.

\paragraph{What the SVD is actually doing.}
If the directions are not used, why does the SVD frame work out of the box while raw random frames need a correction at all? The answer is scale bookkeeping. A transformer has hundreds of target modules whose weight magnitudes differ widely, yet the pipeline drives all of them with a single $\sigma$; something must translate that one global setting into an appropriate per-layer perturbation size. In the SVD frame this translation happens automatically, since singular vectors have unit norm, so the size of $\Delta W$ is set by the layer's own singular values, making the perturbation proportional to the layer's scale by construction. A raw Gaussian frame has no such proportionality, and its $\Delta W$ norm grows with layer size, so a $\sigma$ that is safe for a small layer destabilizes a large one. Restoring the proportionality is exactly what $\alpha$ does; this is the entire content of the R2/R3 protocol above. The correction is not a detail but the load-bearing ingredient, because without it the random variants collapse to zero accuracy at every vote size and every $\sigma$ across two decades (Appendix~D). The SVD's role in this pipeline is therefore not to provide directions but to \emph{calibrate per-layer perturbation scale automatically}, a function that survives being replaced by a single number.

\section{Perturbation Norm, the Only Live Factor}
\label{sec:norm}

\paragraph{Everything else is dead.}
Putting the eliminations together, dimension does not matter (Section~\ref{sec:dimension}), and neither the frame nor its content matters (Section~\ref{sec:subspace}). The only intervention that reliably changes outcomes is the perturbation scale $\sigma$. Gradients do not change this picture. Group Relative Policy Optimization (GRPO) \citep{shao2024deepseekmathpushinglimitsmathematical} in the same $v_\text{dim}{=}12$ space yields no meaningful gain (${+}0.4$ points over base, versus ${+}5.7$ for random search in the same space), and letting gradients \emph{update} the perturbation, as in MeZO, diverges in two of three seeds. Across three seeds the final perturbation norm grows to 0.27, 1.08, and 1.66, with accuracy falling monotonically from .461 to .401 to .160 (details in Appendix~E). The MeZO failure is direct evidence for the norm account, since noisy gradients push the perturbation out of its safe range, and performance dies with the norm.

\paragraph{$\sigma$ has a sharp safe window.}
Unlike every dial eliminated so far, $\sigma$ has unmistakable structure. Above an upper threshold, performance collapses abruptly rather than degrading gracefully. There is no matching lower cliff; small perturbations leave candidates near-clones, giving selection nothing to work with, but do not damage the model (Appendix~C). The collapse curves of Figure~\ref{fig:collapse} show this upper edge, where the SVD frame's mean candidate score falls from .369 to .003 within one step of the $\sigma$ grid. Below that edge lies a usable band, and per-task optima $\sigma^{*}$ concentrate inside it, from .019 on GSM8K to .100 on USPTO-50K (Table~\ref{tab:sigma}).

\begin{table}[!t]
\centering
\small
\setlength{\tabcolsep}{3.5pt}
\begin{tabular}{lccccccc}
\toprule
 & \textbf{GSM} & \textbf{MBPP} & \textbf{Cnt} & \textbf{MATH} & \textbf{Oly} & \textbf{ROC} & \textbf{USP} \\
\midrule
$\sigma^{*}$ & .019 & .026 & .036 & .036 & .036 & .057 & .100 \\
\bottomrule
\end{tabular}
\caption{Per-task optimal $\sigma^{*}$, task mean over models.}
\label{tab:sigma}
\end{table}

\begin{figure}[!t]
\centering
\includegraphics[width=0.95\columnwidth]{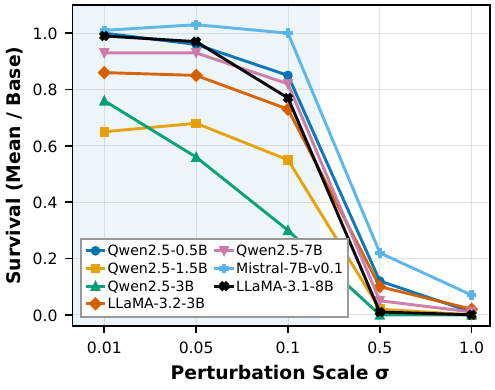}
\caption{Survival curves (mean/base) across seven models (0.5B--8B, three families). The shaded region runs from the smallest $\sigma$ tested to $\sigma_{1/2}{=}0.16$, the mean of the per-model half-survival crossings, which themselves span 0.06 to 0.28 while in-window profiles differ.}
\label{fig:survival}
\end{figure}

\paragraph{The window closes within a factor of five across models.}
Across all seven models (0.5B--8B, three families), survival falls off over the same stretch of the $\sigma$ axis (Figure~\ref{fig:survival}), even though the in-window profiles differ by model. To state this without relying on the eye, we locate each model's upper edge as the $\sigma$ at which survival crosses one half, interpolated in $\log \sigma$. The seven edges average $\sigma_{1/2}{=}0.16$ and span 0.06 to 0.28, a factor of under five. Every crossing is interpolated inside one grid gap, so it bounds rather than resolves the variation. The spread stays under one of the two decades the sweep covers and is not explained by model size, since the smallest and one of the largest models sit at 0.22 and 0.20 while the 3B model sits lowest at 0.06. We therefore take $\sigma{=}0.05$, below the lowest edge we measured. The margin to the 3B edge is only 1.2, so the criterion is conservative rather than comfortable.

\paragraph{And its interior is flat.}
Within the window, precision does not pay. Fixing $\sigma$ at the window midpoint (one value, no per-cell search) costs a median of 0.1 points relative to each cell's own optimum. The selection statistics say the same thing from the other side. The top-$K$ survivors that the pipeline ends up voting with are drawn nearly uniformly from the in-window $\sigma$ values (31\%/38\%/29\% from $\sigma{=}.01/.05/.1$) and almost never from outside it (2\% at .5 and none at 1.0). Any in-window value works. Tasks do have mildly different preferred points, but not knowing them costs almost nothing.

\section{Discussion}
\label{sec:discussion}

\paragraph{Practical prescription.}
The guidance is two lines. Use any frame, since the only ingredient that matters is a per-layer scale correction, which the SVD supplies automatically and a random frame recovers with one scalar. Fix $\sigma$ at 0.05, below the lowest edge we measured, since the same value transfers across models.

\paragraph{A geometric reading.}
Our findings mirror the geometry of diffusion latent spaces \citep{park2023understanding}, where singular values encode how strongly the output responds when a direction is perturbed, and our search consumes only that magnitude channel. The early collapse of the SVD frame (Figure~\ref{fig:collapse}) fits the same reading, since top singular directions are maximum-sensitivity directions and at equal norm perturb the function hardest (Appendix~I).

\paragraph{Relation to prior findings.}
The expert-density observation of Neural Thickets \citep{gan2026thickets} reproduces in our grid, but search dimension turns out not to be the operative factor. The contrast with LoRA-XS \citep{balazy2025loraxs} plausibly traces to gradients, since with a \emph{trained} inner matrix the SVD frame edges out a random one in our replication, by a margin within seed noise (Appendix~F), whereas under gradient-free search only scale survives. LottaLoRA \citep{hazan2026littlerankgoeslong} reports content-independence of frozen scaffolds in the gradient regime; we establish the same proposition in the gradient-free regime, and add the mechanism, norm calibration, together with direct subspace-distance measurements.

\paragraph{Limitations.}
Causal interventions were run on one control cell (Qwen2.5-0.5B / GSM8K, three seeds); the 49-cell grid is single-seed, though pipeline-level seed robustness holds on three representative cells (Appendix~G). All benchmarks are generative tasks with extractable answers, as majority voting requires. Finally, the scoring, selection, and aggregation stages, and comparisons with output-space diversity techniques, are outside our scope.

\section{Conclusion}
\label{sec:conclusion}

In gradient-free perturbation adaptation, survival is decided by the perturbation norm rather than the dimension of the search or the basis that carries it, though where the perturbation is placed is not inert (Appendix~C). The subspace merely calibrates per-layer scale, a function one scalar reproduces, and the norm needs setting only once, since its safe window transfers across sizes and families. The design question narrows from \emph{which subspace to perturb} to \emph{how hard to shake}.

\clearpage
\bibliography{references}

\clearpage
\suppressfloats[t]
\section*{Appendix}
\appendix

\section{Extended Related Work and Method Definitions}
\label{app:extended-related}

Table~\ref{tab:param-counts} organizes the methods of the Related Work section by parameterization, gradient use, and effective search dimension. Our testbed occupies the gradient-free, ultra-low-dimensional cell of weight-space perturbation, which the methods surveyed here leave empty, since the low-dimensional adapters (LoRA-XS, VeRA, TinyLoRA) all keep a gradient backend, while the gradient-free methods (PEP, RandOpt) all perturb the full weight tensor. Gradient-free search in other spaces, prompts and activations among them, is a separate line and is not compared here.

\begin{table}[tbp]
\centering
\small
\resizebox{\columnwidth}{!}{%
\begin{tabular}{lccc}
\toprule
\textbf{Method} & \textbf{Grad.} & \textbf{Per-module count} & \textbf{Order} \\
\midrule
Full fine-tuning        & yes & $mn$      & $\sim$$10^9$ \\
LoRA / PiSSA           & yes & $r(m{+}n)$ & $\sim$$10^7$ \\
LoRA-XS                 & yes & $r^2$     & $\sim$$10^3$ \\
VeRA                    & yes & $m{+}r$   & $\sim$$10^5$ \\
Subspace + GRPO         & yes & $v_\text{dim}$ tied & $\sim$$10^1$--$10^3$ \\
RandOpt                 & no  & $mn$      & $\sim$$10^9$ \\
Ours (testbed)          & no  & $v_\text{dim}$ tied & $\sim$$10^1$--$10^3$ \\
\bottomrule
\end{tabular}}
\caption{Trainable parameter / effective search dimension per target module ($W \in \mathbb{R}^{m\times n}$, rank $r$; Order is the model-wide order of magnitude).}
\label{tab:param-counts}
\end{table}

\section{Per-Model Search Dimensions}
\label{app:vdim}

Table~\ref{tab:vdim} lists the $v_\text{dim}$ configurations per model. The tiny configuration is 12 or 16 scalars depending on the layer count, and is nearly invariant to model scale.

\begin{table}[tbp]
\centering
\small
\begin{tabular}{lccccc}
\toprule
\textbf{Model} & \textbf{utiny} & \textbf{tiny} & \textbf{small} & \textbf{medium} & \textbf{large} \\
\midrule
Qwen2.5-0.5B    & 4 & 12 & 40 & 192 & 768  \\
Qwen2.5-1.5B    & 4 & 12 & 48 & 224 & 896  \\
Qwen2.5-3B      & 4 & 16 & 64 & 288 & 1152 \\
LLaMA-3.2-3B    & 4 & 12 & 48 & 224 & 896  \\
Qwen2.5-7B      & 4 & 12 & 48 & 224 & 896  \\
Mistral-7B-v0.1 & 4 & 16 & 56 & 256 & 1024 \\
LLaMA-3.1-8B    & 4 & 16 & 56 & 256 & 1024 \\
\bottomrule
\end{tabular}
\caption{Per-model $v_\text{dim}$ across the five capacity configurations.}
\label{tab:vdim}
\end{table}

\section{Hyperparameter Sweep Details}
\label{app:sweep}

\paragraph{One-axis-at-a-time sweep.}
Table~\ref{tab:one-axis} sweeps each numeric dial around the control-cell baseline. All twelve variants land in a narrow band around Base; no numeric axis dominates. The two $\sigma$ rows need care, because they shift the entire search grid by a factor of ten rather than fixing one value, and the pipeline still selects over whatever grid it is given. O12 retains only $\sigma{=}0.1$ inside the window, where O12's own mean candidate score is .390 against .011 or below at every larger value, and selection recovers .471 from that single column. The collapse reported in the norm section of the main paper is what happens when $\sigma$ is held at one misspecified value, which is a different experiment. The low grid also shows that there is no matching lower cliff. At $\sigma$ of .001 and .005 the mean candidate score holds at .517 and .516, so weak perturbations do not damage the model, they merely stop separating candidates.

\begin{table}[tbp]
\centering
\small
\begin{tabular}{llll}
\toprule
\textbf{ID} & \textbf{Axis} & \textbf{Value} & \textbf{Acc.} \\
\midrule
Base & --    & --    & .491 \\
O1   & $K$   & +200  & .476 $\pm$ .024 \\
O2   & $N$   & 100   & .492 $\pm$ .015 \\
O3   & $N$   & 500   & .471 $\pm$ .031 \\
O4   & $N$   & 2000  & .464 $\pm$ .054 \\
O5   & $n_\text{fit}$ & 50  & .461 $\pm$ .052 \\
O6   & $n_\text{fit}$ & 100 & .455 $\pm$ .056 \\
O7   & $n_\text{fit}$ & 500 & .458 $\pm$ .013 \\
O8   & $r$   & 1     & .461 $\pm$ .023 \\
O9   & $r$   & 4     & .461 $\pm$ .022 \\
O10  & $r$   & 8     & .452 $\pm$ .049 \\
O11  & $\sigma$ & grid $\times 0.1$ & .444 $\pm$ .048 \\
O12  & $\sigma$ & grid $\times 10$  & .471 $\pm$ .071 \\
\bottomrule
\end{tabular}
\caption{One-axis-at-a-time sweep on the control cell (three seeds).}
\label{tab:one-axis}
\end{table}

\paragraph{Per-$K$ breakdown.}
Table~\ref{tab:per-k} breaks the sweep down by vote size. A $K$ plateau forms at $K \approx 50$ (O1, the only configuration extended to $K{=}200$, stays flat at .474), and above $K{=}40$ every configuration varies by at most 1.3 points, so best-$K$ reporting is not an ensemble-size artifact.

\begin{table}[tbp]
\centering
\scriptsize
\setlength{\tabcolsep}{3pt}
\resizebox{\columnwidth}{!}{%
\begin{tabular}{lcccccc}
\toprule
\textbf{ID} & \textbf{$K{=}1$} & \textbf{$K{=}10$} & \textbf{$K{=}40$} & \textbf{$K{=}50$} & \textbf{$K{=}100$} & \textbf{$K{=}200$} \\
\midrule
Base & .467 & .478 & .489 & .485 & .491 & -- \\
O1  & .417 $\pm$ .037 & .460 $\pm$ .033 & .474 $\pm$ .024 & .474 $\pm$ .023 & .472 $\pm$ .021 & .474 \\
O2  & .418 $\pm$ .038 & .459 $\pm$ .025 & .479 $\pm$ .016 & .489 $\pm$ .014 & .492 $\pm$ .015 & -- \\
O3  & .415 $\pm$ .039 & .457 $\pm$ .039 & .462 $\pm$ .038 & .466 $\pm$ .038 & .468 $\pm$ .027 & -- \\
O4  & .410 $\pm$ .073 & .456 $\pm$ .050 & .462 $\pm$ .052 & .461 $\pm$ .056 & .460 $\pm$ .060 & -- \\
O5  & .409 $\pm$ .065 & .453 $\pm$ .051 & .453 $\pm$ .054 & .458 $\pm$ .054 & .461 $\pm$ .051 & -- \\
O6  & .400 $\pm$ .049 & .445 $\pm$ .054 & .450 $\pm$ .053 & .454 $\pm$ .056 & .451 $\pm$ .052 & -- \\
O7  & .410 $\pm$ .054 & .452 $\pm$ .015 & .456 $\pm$ .013 & .455 $\pm$ .011 & .448 $\pm$ .023 & -- \\
O8  & .384 $\pm$ .061 & .441 $\pm$ .041 & .458 $\pm$ .025 & .459 $\pm$ .022 & .452 $\pm$ .028 & -- \\
O9  & .396 $\pm$ .063 & .447 $\pm$ .029 & .461 $\pm$ .022 & .456 $\pm$ .025 & .453 $\pm$ .032 & -- \\
O10 & .380 $\pm$ .060 & .440 $\pm$ .052 & .451 $\pm$ .048 & .450 $\pm$ .051 & .452 $\pm$ .049 & -- \\
O11 & .392 $\pm$ .062 & .438 $\pm$ .042 & .437 $\pm$ .058 & .435 $\pm$ .056 & .434 $\pm$ .051 & -- \\
O12 & .335 $\pm$ .078 & .432 $\pm$ .076 & .467 $\pm$ .068 & .471 $\pm$ .072 & .469 $\pm$ .069 & -- \\
\bottomrule
\end{tabular}}
\caption{Per-$K$ ensemble accuracy for the sweep configurations (GSM8K, three seeds). The Base row is the single reference run; the three-seed rerun of that configuration reads .484 $\pm$ .007 at $K{=}40$ in the frame ablation of the main paper, one seed deviation away.}
\label{tab:per-k}
\end{table}

\paragraph{Layer group and module type.}
Restricting the perturbation to a layer group or a module type also reduces $v_\text{dim}$, so these variants confound location with parameter count. All eleven variants land between .423 and .452, from 3.9 to 6.8 points below the all-layer baseline (Table~\ref{tab:layer-module}), and the gap exceeds what the capacity sweep of the dimension section of the main paper attributes to the reduction in $v_\text{dim}$ alone. Where to perturb is therefore not fully inert, though the effect stays smaller than a misspecified $\sigma$. Mild location trends do exist. The late group leads in two of three module budgets, and MLP-only exceeds attention-only by 1.3--1.7 points in mid and late groups, consistent with AdaLoRA's preferential budget allocation to upper layers and feed-forward modules \citep{zhang2023adalora}.

\begin{table}[tbp]
\centering
\small
\begin{tabular}{llccc}
\toprule
\textbf{ID} & \textbf{Layers} & \textbf{Modules} & \textbf{$v_\text{dim}$} & \textbf{Acc.} \\
\midrule
Base & all   & both      & 12 & .491 \\
O1   & all   & attn-only & 6  & .444 $\pm$ .009 \\
O2   & all   & MLP-only  & 6  & .448 $\pm$ .018 \\
O3   & early & both      & 4  & .445 $\pm$ .020 \\
O4   & early & attn-only & 2  & .431 $\pm$ .005 \\
O5   & early & MLP-only  & 2  & .431 $\pm$ .014 \\
O6   & mid   & both      & 4  & .444 $\pm$ .012 \\
O7   & mid   & attn-only & 2  & .423 $\pm$ .003 \\
O8   & mid   & MLP-only  & 2  & .436 $\pm$ .006 \\
O9   & late  & both      & 4  & .452 $\pm$ .002 \\
O10  & late  & attn-only & 2  & .430 $\pm$ .017 \\
O11  & late  & MLP-only  & 2  & .447 $\pm$ .008 \\
\bottomrule
\end{tabular}
\caption{Layer group $\times$ module type ablation. Restricting location also lowers $v_\text{dim}$, so the two are confounded here.}
\label{tab:layer-module}
\end{table}

\section{Frame Ablation Details}
\label{app:frame-details}

This appendix specifies the four frame variants of the subspace section of the main paper and shows that the norm correction is what keeps the random variants alive.

\paragraph{Formal definitions.}
Let $W = U S V^\top$ be the top-$r$ SVD and $v \in \mathbb{R}^{v_\text{dim}}$ the perturbed scalar group ($r{=}2$, $v_\text{dim}{=}12$, and $n_\text{tie}{=}10$ consecutive layers of a given module type sharing one scalar, across the 96 target modules vLLM exposes after fusing qkv and gate-up; the same weights count as 168 modules before fusion). All variants share the identical scalar count; only the lift from $v$ to $\Delta W$ differs (Table~\ref{tab:variant-defs}). For R2, $P$ is $n \times n$ rather than $r \times r$, because the low-rank structure is part of the SVD prior, so the ablation must not reintroduce it through $P$.

\paragraph{The norm correction $\alpha$.}
A random lift and the SVD lift it replaces carry different Frobenius norms, so R2 and R3 receive one scalar per module that equalizes them at $|v|{=}1$,
\begin{equation}
\alpha_i = \|U_i S_i V_i^\top\|_F \,\big/\, \|M_i P_i\|_F ,
\end{equation}
with $U_i S_i V_i^\top$ the rank-$r$ reconstruction of module $i$. This is the only quantity matched across variants, and Tables~\ref{tab:uncorrected-k} and~\ref{tab:uncorrected-sigma} show what happens without it. The global-$\alpha$ control of Figure~\ref{fig:collapse} replaces every $\alpha_i$ by the single ratio of the two module-averaged norms, so the overall scale is matched while the per-layer profile is not.

\begin{table}[!htb]
\centering
\small
\setlength{\tabcolsep}{4pt}
\begin{tabular}{lcccl}
\toprule
\textbf{Variant} & \textbf{$\Delta W$} & \textbf{SVD} & \textbf{$vP$} & \textbf{Rank / Transform} \\
\midrule
Base & $US(vP)V^\top$ & \checkmark & \checkmark & $r{=}2$, top-$r$ SVD \\
R1   & $v \cdot USV^\top$ & \checkmark & -- & $r{=}2$, top-$r$ SVD \\
R2   & $v \cdot MP$ & -- & \checkmark & full-$n$, raw $\mathcal N(0,1)$ \\
R3   & $v \cdot M$  & -- & -- & full-$n$, raw $\mathcal N(0,1)$ \\
\bottomrule
\end{tabular}
\caption{Definitions of the four frame variants.}
\label{tab:variant-defs}
\end{table}

\paragraph{Norm correction is the essential ingredient.}
Without the per-module scale correction $\alpha$ (computed from weight-norm ratios), the raw random frames are unrecoverable, with accuracy at or near .000 across the whole vote-size range (Table~\ref{tab:uncorrected-k}) and train scores at noise across two decades of $\sigma$ (Table~\ref{tab:uncorrected-sigma}). This is the direct evidence for the claim of the subspace section of the main paper that scale calibration, not direction, is what the frame must supply.

\begin{table}[!htb]
\centering
\small
\setlength{\tabcolsep}{3pt}
\resizebox{\columnwidth}{!}{%
\begin{tabular}{lccccc}
\toprule
\textbf{Variant} & \textbf{$K{=}1$} & \textbf{$K{=}10$} & \textbf{$K{=}40$} & \textbf{$K{=}50$} & \textbf{$K{=}100$} \\
\midrule
Base & .467 & .478 & .489 & .485 & .491 \\
R1 (no $vP$) & .452 $\pm$ .006 & .485 $\pm$ .014 & .479 $\pm$ .006 & .480 $\pm$ .004 & .476 $\pm$ .012 \\
R2 raw (no $\alpha$) & .004 $\pm$ .001 & .001 $\pm$ .001 & .000 & .000 & .000 \\
R3 raw (no $\alpha$) & .005 $\pm$ .001 & .003 $\pm$ .001 & .000 & .000 & .000 \\
\bottomrule
\end{tabular}}
\caption{Per-$K$ accuracy \emph{without} norm correction. Raw random frames collapse at every $K$.}
\label{tab:uncorrected-k}
\end{table}

\begin{table}[!htb]
\centering
\small
\setlength{\tabcolsep}{3pt}
\resizebox{\columnwidth}{!}{%
\begin{tabular}{lccccc}
\toprule
\textbf{Variant} & \textbf{$\sigma{=}.01$} & \textbf{.05} & \textbf{.1} & \textbf{.5} & \textbf{1.0} \\
\midrule
Base & .588 $\pm$ .018 & .555 $\pm$ .033 & .492 $\pm$ .074 & .018 $\pm$ .058 & .001 $\pm$ .002 \\
R1   & .582 $\pm$ .018 & .561 $\pm$ .034 & .506 $\pm$ .065 & .031 $\pm$ .075 & .001 $\pm$ .003 \\
R2 raw & .002 $\pm$ .003 & .001 $\pm$ .003 & .001 $\pm$ .003 & .002 $\pm$ .003 & .001 $\pm$ .003 \\
R3 raw & .004 $\pm$ .005 & .001 $\pm$ .003 & .001 $\pm$ .003 & .001 $\pm$ .003 & .001 $\pm$ .002 \\
\bottomrule
\end{tabular}}
\caption{Sampling-phase train accuracy \emph{without} norm correction. Raw frames stay at noise across two decades of $\sigma$. This Base row and the collapse figure of the subspace section come from different runs of the same configuration, which is why their $\sigma{=}0.1$ values differ.}
\label{tab:uncorrected-sigma}
\end{table}

\section{Gradient-Based Baselines}
\label{app:gradient-baselines}

Table~\ref{tab:grpo-detail} details the gradient comparisons of the norm section of the main paper; Table~\ref{tab:grpo-hp} lists GRPO hyperparameters. In the MeZO runs, the final perturbation norm tracks the accuracy loss monotonically, since the update pushes the norm out of the safe window.

\begin{table}[tbp]
\centering
\small
\resizebox{\columnwidth}{!}{%
\begin{tabular}{lccc}
\toprule
\textbf{Method} & \textbf{Base} & \textbf{Acc.} & \textbf{$\Delta$} \\
\midrule
Subspace + GRPO (3 seeds) & .4215 & .4258 $\pm$ .0065 & +0.4 \\
Subspace tiny ($K{=}1$)   & .4337 & .4670 & +3.3 \\
Subspace tiny ($K{=}100$) & .4337 & .4905 & +5.7 \\
Full-Weight (FW) RandOpt ($K{=}50$) & .4337 & .5250 & +9.1 \\
\midrule
\multicolumn{4}{l}{\textbf{MeZO} (base .4443; final norm of $v \to$ test acc.)} \\
seed 42 & \multicolumn{3}{l}{norm 0.27 $\to$ .461 (+1.7)} \\
seed 1  & \multicolumn{3}{l}{norm 1.08 $\to$ .401 ($-$4.3)} \\
seed 2  & \multicolumn{3}{l}{norm 1.66 $\to$ .160 ($-$28.4)} \\
\bottomrule
\end{tabular}}
\caption{Gradient-based alternatives in the same space. GRPO barely improves over base; MeZO's accuracy falls monotonically with its final perturbation norm. Each method evaluates the unperturbed model through its own harness, so the base column spans .4215 to .4443 on the same cell. The comparable quantity is therefore $\Delta$ within a row, and the 1.2-point harness spread is smaller than the contrast the table is used for.}
\label{tab:grpo-detail}
\end{table}

\begin{table}[tbp]
\centering
\small
\begin{tabular}{ll}
\toprule
\textbf{Hyperparameter} & \textbf{Value} \\
\midrule
lr & $2\times10^{-4}$ \\
num\_epochs & 25 \\
batch\_size & 64 \\
group\_size $G$ & 8 \\
clip\_eps & 0.2 \\
kl\_beta & 0.001 \\
optimizer & Adam \\
seed & 42 \\
KL estimator & k3 \\
\bottomrule
\end{tabular}
\caption{GRPO hyperparameters.}
\label{tab:grpo-hp}
\end{table}

\section{LoRA-XS Parameterization under Gradient Training}
\label{app:loraxs}

When the same frames are \emph{trained} rather than searched, the SVD information is consumed, and the SVD frame outperforms a random orthogonal frame by 0.7 points (Table~\ref{tab:loraxs}). The contrast with the gradient-free results of the subspace section of the main paper localizes the difference to the presence of gradients.

\begin{table}[tbp]
\centering
\small
\begin{tabular}{lcc}
\toprule
\textbf{Parameterization} & \textbf{$v_\text{dim}$} & \textbf{Best ens.\ acc.} \\
\midrule
Base (unperturbed)              & --  & .435 \\
LoRA-XS, SVD frame              & 48  & .480 \\
LoRA-XS, random orthogonal frame & 48 & .473 \\
\bottomrule
\end{tabular}
\caption{LoRA-XS comparison under gradient training.}
\label{tab:loraxs}
\end{table}

\section{Seed Robustness}
\label{app:seeds}

Repeating the full pipeline with three seeds on three representative cells preserves all conclusions (Table~\ref{tab:seeds}), mitigating the single-seed limitation of the 49-cell grid.

\begin{table}[tbp]
\centering
\small
\setlength{\tabcolsep}{4pt}
\begin{tabular}{lcccc}
\toprule
\textbf{Cell} & \textbf{s42} & \textbf{s1} & \textbf{s2} & \textbf{Mean $\pm$ std} \\
\midrule
Qwen2.5-0.5B / GSM8K    & .488 & .491 & .488 & .489 $\pm$ .002 \\
Qwen2.5-0.5B / MATH-500 & .383 & .367 & .357 & .369 $\pm$ .013 \\
Qwen2.5-3B / Countdown  & .423 & .470 & .453 & .449 $\pm$ .024 \\
\bottomrule
\end{tabular}
\caption{Pipeline-level seed robustness on representative cells ($K{=}100$).}
\label{tab:seeds}
\end{table}

\section{Implementation Details}
\label{app:implementation}

Models are served in bf16 with vLLM \citep{kwon2023vllm}, up to four engines per GPU using fractional memory allocation. All evaluations use each model's instruct chat template and greedy decoding. Scoring is binary exact-match for math and code, partial credit for Countdown, and classification accuracy for ROCStories and USPTO-50K. Table~\ref{tab:cost} reports the cost anatomy of one adaptation run. Candidate scoring dominates at 99.5\%, so neither dimension nor frame has a lever on cost. Prompt templates for each task follow the dataset handlers in our released code.

\begin{table}[tbp]
\centering
\small
\setlength{\tabcolsep}{4pt}
\begin{tabular}{lcc}
\toprule
 & \textbf{Subspace} & \textbf{Full-Weight} \\
\midrule
Search dimension & 12 & $4.9\times10^{8}$ \\
One-time init (SVD) & 5.0 $\pm$ 0.0 s & -- \\
Apply one perturbation & 28.4 $\pm$ 1.6 ms & 13.4 $\pm$ 0.4 ms \\
Score one candidate (200) & 11.6 $\pm$ 0.6 s & 11.4 $\pm$ 0.0 s \\
Adaptation total ($N{=}1000$) & 3.23 $\pm$ 0.16 h & 3.17 $\pm$ 0.01 h \\
Share of scoring & 99.7\% & 99.9\% \\
Peak GPU memory & 3.63 GiB & 3.23 GiB \\
\bottomrule
\end{tabular}
\caption{Cost anatomy of one adaptation run.}
\label{tab:cost}
\end{table}
\section{Representation-Level Analysis (SAE)}
\label{app:sae}

\begin{figure*}[p]
\centering
\includegraphics[width=0.94\textwidth]{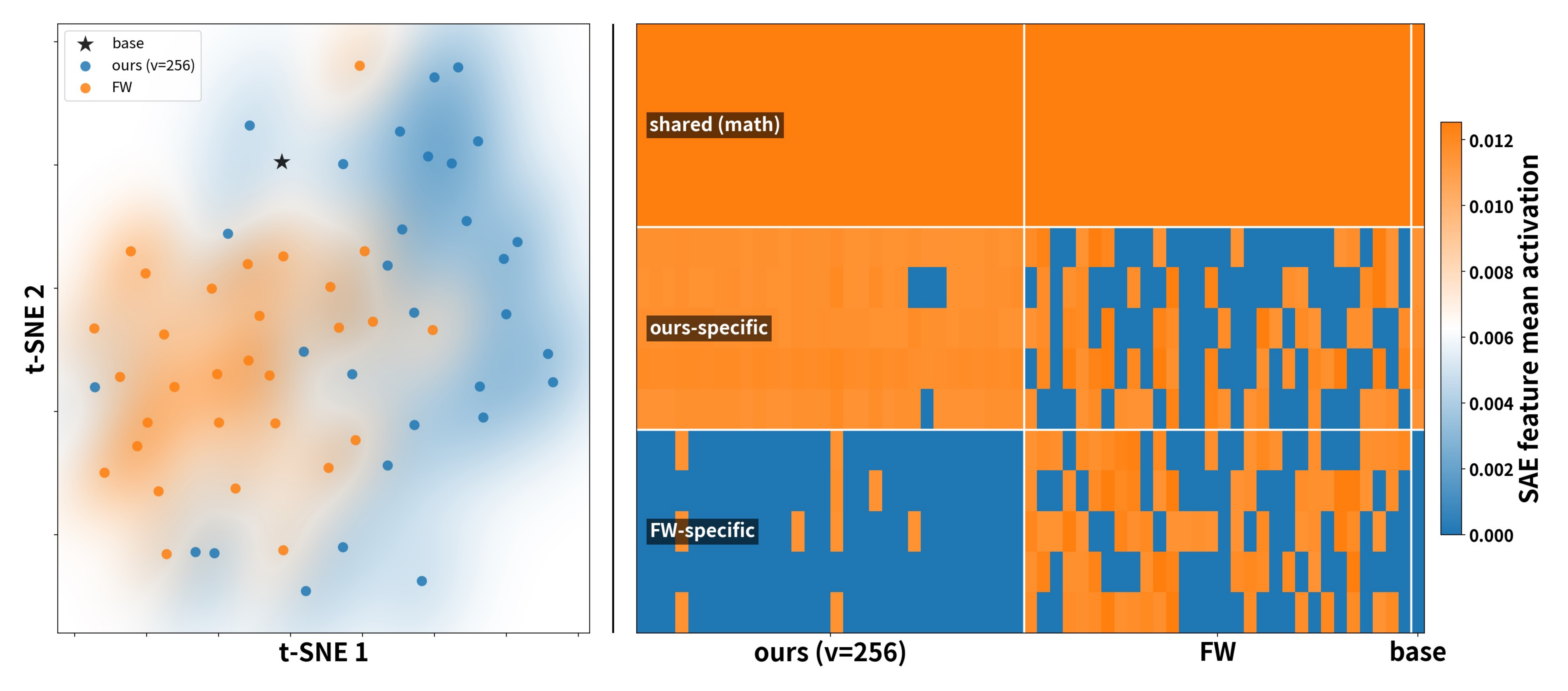}
\caption{Left: t-SNE of perturbed-model representations around base. Right: SAE feature activation heatmap; subspace perturbations stay near the base signature while full-weight perturbations activate FW-specific features.}
\label{fig:sae}
\end{figure*}

Measured with a sparse autoencoder (SAE) dictionary at layer 21 on the \emph{medium} capacity over 100 held-out problems, the representation deviation of subspace perturbations from base is 13--19$\times$ smaller than that of full-weight perturbations (Table~\ref{tab:sae}, Figure~\ref{fig:sae}), supporting the geometric reading of the discussion of the main paper that safe-window perturbations stay in the local neighborhood of the base representation.

\begin{table}[tbp]
\centering
\small
\setlength{\tabcolsep}{4pt}
\resizebox{\columnwidth}{!}{%
\begin{tabular}{lccc}
\toprule
\textbf{Metric} & \textbf{Subspace} & \textbf{FW} & \textbf{Ratio} \\
\midrule
Answer change ratio (Hamming) & 15.5\% & 19.7\% & 1.27$\times$ \\
Accuracy gap from base & $-0.5$ pts & $-1.77$ pts & 3.5$\times$ \\
SAE features off-base, $>1$ s.d. & 0.35\% & 4.61\% & 13.2$\times$ \\
SAE features off-base, $>2$ s.d. & 0.11\% & 2.11\% & 18.8$\times$ \\
SAE cosine distance to base & 0.0028 & 0.0173 & 6.2$\times$ \\
\bottomrule
\end{tabular}}
\caption{Representation-level deviation from base (MATH-500, \emph{medium} capacity, layer 21). Accuracies are means over the sampled perturbed models rather than the top-$K$ ensemble, which is why both arms sit below base here while the voted ensembles of the main grid sit above it.}
\label{tab:sae}
\end{table}

\section{Performance Distributions of Adapted Models}
\label{app:distributions}

\begin{figure*}[p]
\centering
\includegraphics[width=0.82\textwidth]{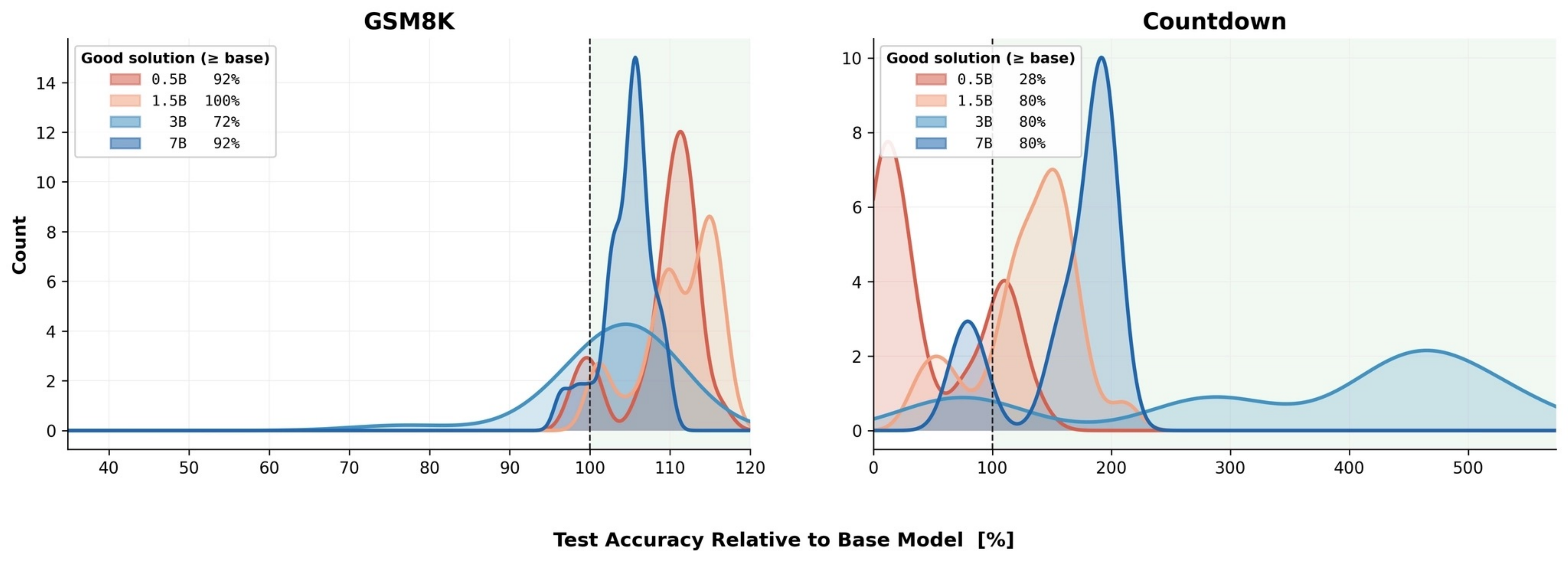}
\caption{Test-set performance distributions of adapted models relative to base on GSM8K (left) and Countdown (right). The shaded region marks ensembles at or above base.}
\label{fig:histogram}
\end{figure*}

Figure~\ref{fig:histogram} reports the test-set accuracy distribution of pipeline outputs relative to each model's base on GSM8K and Countdown, as descriptive statistics only. Most cells sit at or above base, confirming that the grid-level results are not driven by a few lucky tails.

\section{Weight-Space Landscapes}
\label{app:landscape}

\begin{figure*}[p]
\centering
\includegraphics[width=0.86\textwidth]{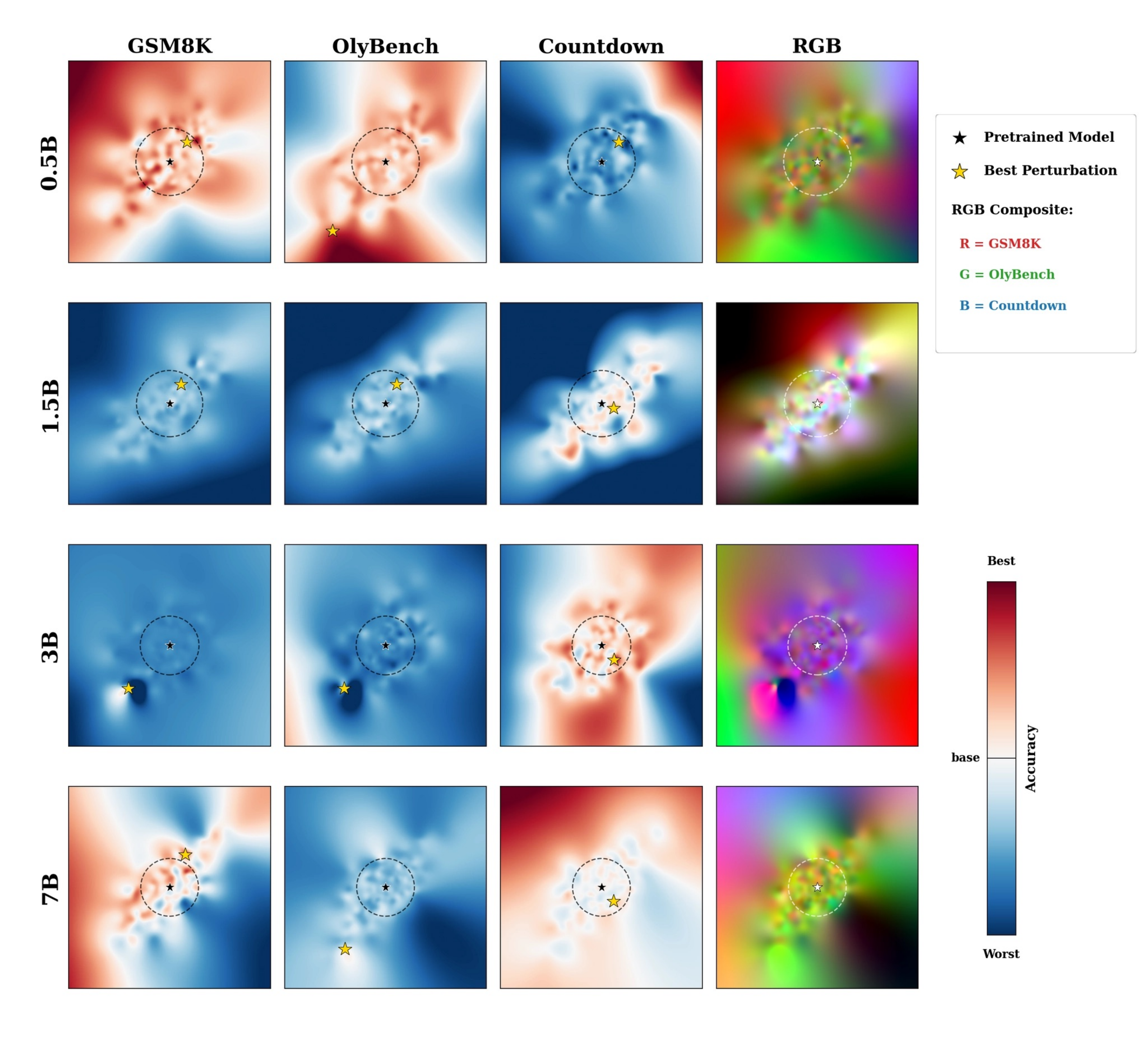}
\caption{Accuracy landscapes in weight space (0.5B--7B) under 200 random weight perturbations projected to 2D. Color shows relative accuracy change; the dashed circle marks mean perturbation distance and stars mark best perturbations. The last column maps three tasks to RGB channels.}
\label{fig:landscape}
\end{figure*}

Figure~\ref{fig:landscape} visualizes 2D-projected accuracy landscapes around each base model under 200 random full-weight perturbations. Larger models display a denser improved region near base, consistent with the Neural Thickets density argument \citep{gan2026thickets}; the best perturbations for different tasks rarely coincide, which illustrates why single-candidate cells are noisy.

\end{document}